\documentclass[sigconf]{acmart}

\usepackage{balance}
\usepackage{subcaption}

\def\BibTeX{{\rm B\kern-.05em{\sc i\kern-.025em b}\kern-.08emT\kern-.1667em\lower.7ex\hbox{E}\kern-.125emX}}
    
\copyrightyear{2020}
\acmYear{2020}
\setcopyright{rightsretained}
\acmConference[HRI '20 Companion]{Companion of the 2020 ACM/IEEE
International Conference on Human-Robot Interaction}{March 23--26,
2020}{Cambridge, United Kingdom}
\acmBooktitle{Companion of the 2020 ACM/IEEE International Conference
on Human-Robot Interaction (HRI '20 Companion), March 23--26, 2020,
Cambridge, United Kingdom}\acmDOI{10.1145/3371382.3378260}
\acmISBN{978-1-4503-7057-8/20/03}

\newcommand\blfootnote[1]{%
  \begingroup
  \renewcommand\thefootnote{}\footnote{#1}%
  \addtocounter{footnote}{-1}%
  \endgroup
}

\begin{document}

\fancyhead{}

\title{Evaluation of the Handshake Turing Test for anthropomorphic Robots}


\author{Ruth Stock-Homburg, Jan Peters, Katharina Schneider, Vignesh Prasad, Lejla Nukovic}

\affiliation{%
	\institution{Technische Universit\"at Darmstadt, Germany}
}



%
\renewcommand{\shortauthors}{Nukovic et al.}

%
\begin{abstract}
Handshakes are fundamental and common greeting and parting gestures among humans. They are important in shaping first impressions as people tend to associate character traits with a person's handshake. To widen the social acceptability of robots and make a lasting first impression, a good handshaking ability is an important skill for social robots. Therefore, to test the human-likeness of a robot handshake, we propose an initial Turing-like test, primarily for the hardware interface to future AI agents. We evaluate the test on an android robot's hand to determine if it can pass for a human hand. This is an important aspect of Turing tests for motor intelligence where humans have to interact with a physical device rather than a virtual one. We also propose some modifications to the definition of a Turing test for such scenarios taking into account that a human needs to interact with a physical medium.
\end{abstract}

%
%


%

%
\maketitle

\blfootnote{Primary Author's contact: lejla.nukovic@gmail.com\\
Jan Peters is also affiliated to MPI for Intelligent Systems, Tuebingen, Germany\\
}

\vspace{-2.5em}

\section{Introduction}
The original Turing Test for AI agents was limited to a computer interface since there was no hardware involved for testing the agents. In the case of social robots and HRI, a test for motor intelligence would be much more complex especially for android robots hands like the one shown in Fig. \ref{fig:elenoide}. With such interfaces, the way hardware is built must be considered along with the movements. If participants can easily distinguish that they are interacting with a robotic hand, such an interface would fail the Turing test even if the handshaking algorithm is extremely human-like. Another interpretation of the Turing test is for motor intelligence that only tests the motion of the interface interacting with the human. There are only two such experiments for such a Turing-Like handshaking test and were done on 1-D stylus\cite{10.1007/978-3-642-14064-8_29,TowardPerceiving2012}. Though the models for a handshake are well thought, they would need to be redefined when transferred to a more complex telerobotic interface. Moreover, all models concentrate on just the shaking movement during a handshake and ignore other aspects, like the reaching and grasping of the hand, etc., as they are limited with a simple haptic interface \cite{10.1007/978-3-642-14064-8_29,ThreeAlternatives2012,article,TowardPerceiving2012,giannopoulos2011comparison}. Evaluating the human-likeness of a handshake on a more complex android robot yields different challenges. All other aspects of a handshake mentioned above must be considered for a Handshake Turing test, some of which have been evaluated in \cite{knoop2017handshakiness}.

\section{Experiment}

  \begin{figure}
  \centering
  \begin{subfigure}[b]{0.27\textwidth}
                \includegraphics[width=0.9\linewidth]{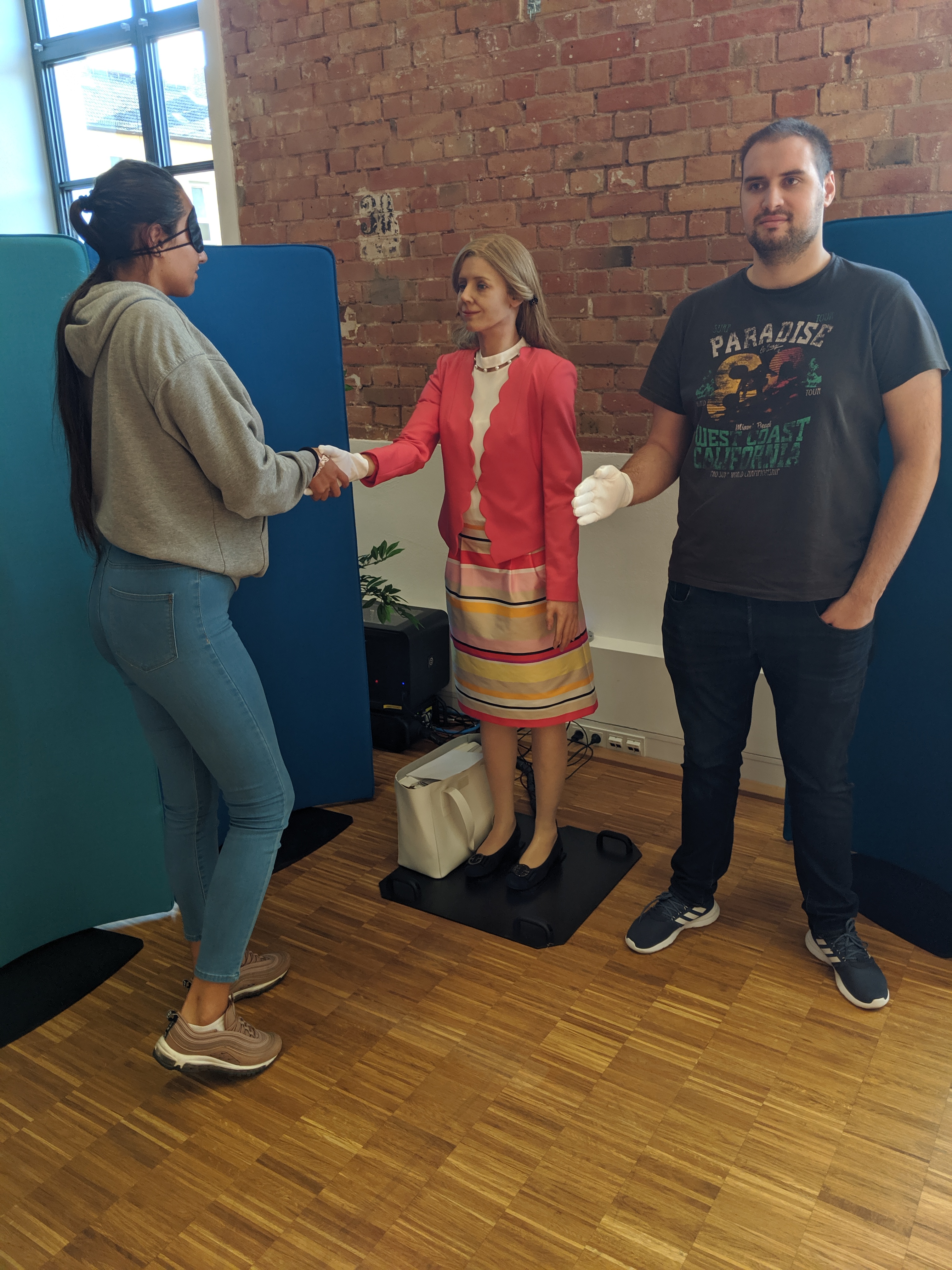}
                \caption{Experiment Scenario}
                \label{fig:exp}
        \end{subfigure}%
        \begin{subfigure}[b]{0.17\textwidth}
                \includegraphics[width=0.9\linewidth]{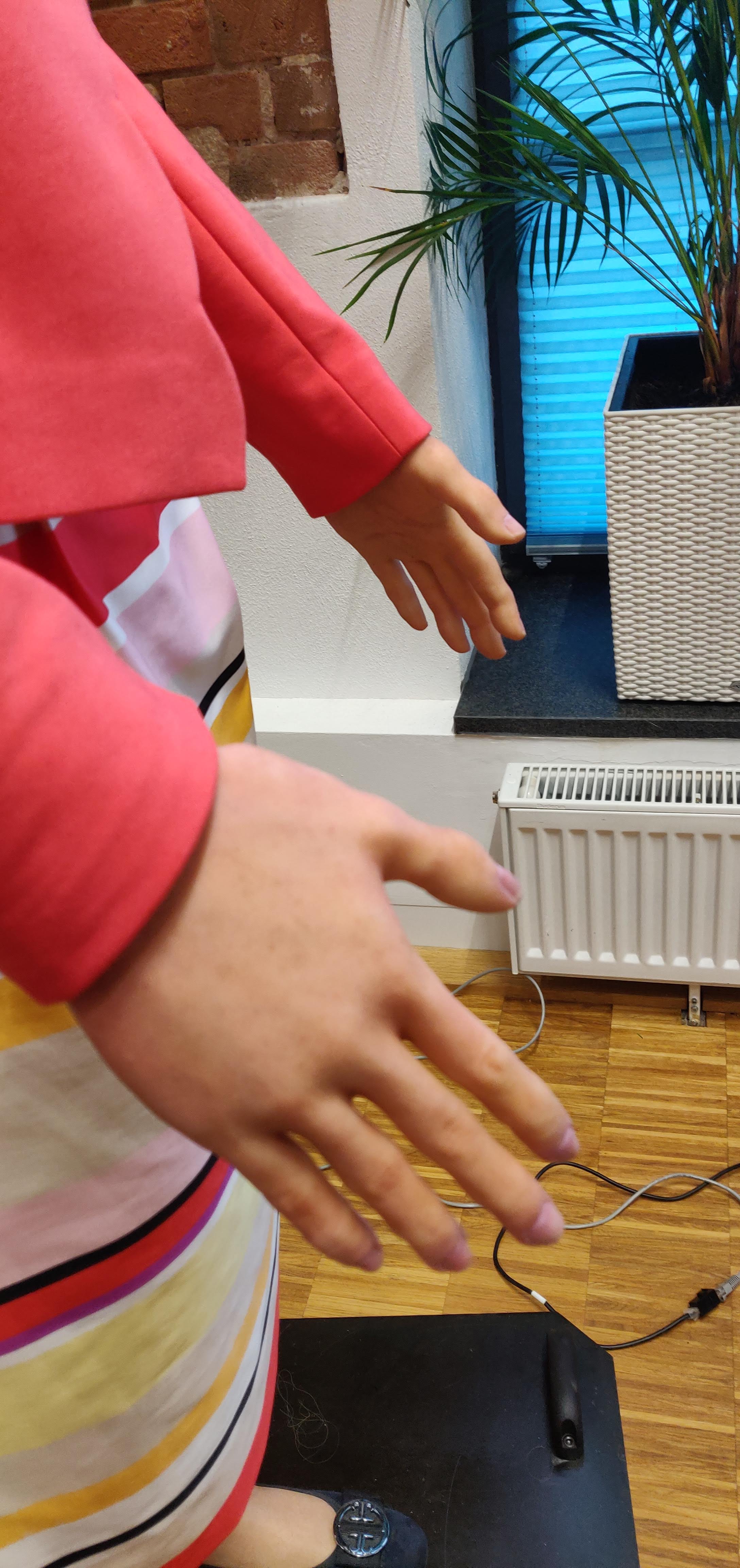}
                \caption{Robot Hand}
                \label{fig:elenoide}
        \end{subfigure}%
        \vspace{-1em}
		
	
	\caption{Setup for the study}
    \label{fig:exp_setup}

 	\vspace{-2em}
\end{figure} 

The aim of this experiment is primarily to explore, whether the android robot's hand shown in Fig. \ref{fig:exp_setup} can be distinguished when felt by a human. In the original Turing test, if participants didn't recognize they were communicating with a machine, then the machine passed the Turing test\cite{10.1093/mind/LIX.236.433}. 
In this study, a similar test is applied to a robot's hand to determine if it can be confused for a human's. This is to determine if such an interface is suitable for further research in physical HRI to solve complicated Turing-like tests, where the robot can effectively convey the feeling of interacting with a human.
\subsection{Experiment Setup}

We use an anthropomorphic android robot designed after a human both in terms of appearance and functionality, to such an extent that the hands have heating pads inside the palms so as to give a feeling of human touch \cite{cabibihan2014illusory}. For this experiment, both the robot hand and a human test subject's hand were covered with a white butler glove as shown in Fig. \ref{fig:exp}. Their arms were lifted from the elbow down in a suitable position so that participants can easily access their hands. Consent to take part in the study with all necessary and legal information is handed out to participants to sign. After signing, they are asked to complete a questionnaire about personal demographic data and technology familiarity. They are informed that the robot and the human subject don’t give any feedback, as the task is to decide, only by feeling the hand, if they are touching the robot hand or the human hand. They are blindfolded and led to one of the hands to feel and decide if it’s a human or robot hand. If they determine that the hand is robot hand regardless if this is true or not they are asked if they find this encounter pleasant or not. Every participant is led two times each to the human hand and the robot hand in a random order. After the testing of hands is done, participants are allowed to take blindfolds off and are immediately asked to fill out a questionnaire about their current emotional state.

\section{Results}
\begin{figure}
	\includegraphics[width=0.45\textwidth]{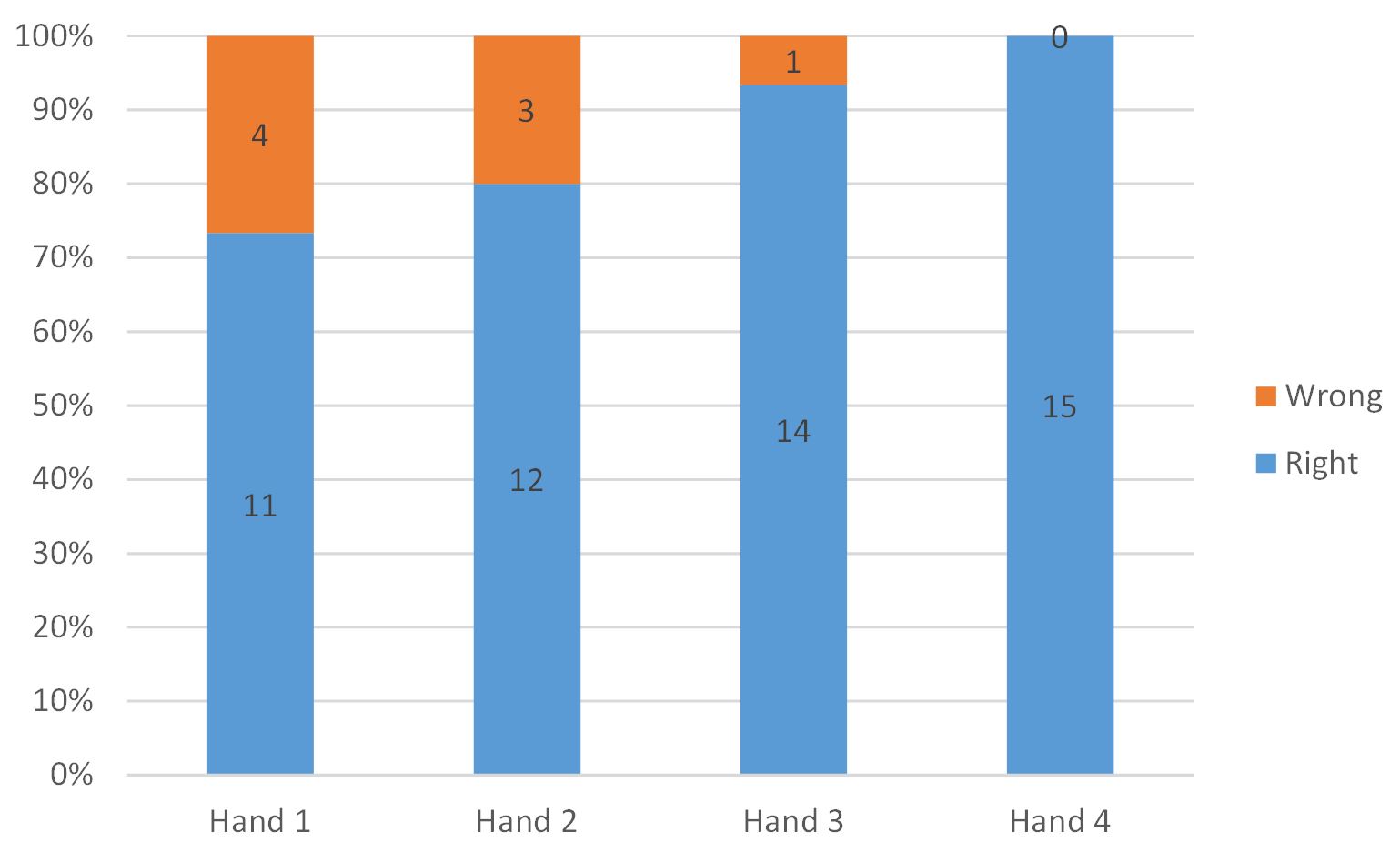}
	\caption{Results of the experiment. (Blue - No. of people who correctly identified the hand, Orange - No. of people who wrongly identified the hand)}
	\label{fig:results}
\end{figure}

As mentioned in Sec. 2.1, each participant was subjected to 2 interactions each with the human and the robot in random order. The results of this study are visualized in Fig. \ref{fig:results} and are explained below.  With the first interaction, 11 of 15 participants $(73 \%)$ guessed right. The other 4 participants $(27 \%)$ mistook the human's hand for a robot's. Interestingly 12 of 15 participants $(80 \%)$ guessed the hand in the second interaction correctly and the other 3 $(20 \%)$ mistook a robot hand for a human hand. The majority of participants (14 of 15) recognized their third interaction correctly. The one participant who was wrong believed that it was a human hand although it was a robot hand. Every participant perceived the true nature of the hand in the fourth interaction. 

Approximately half of the participants $ (57 \%)$ perceived the robot hand as pleasant. On asking the participants how they perceived the robot hand, the majority (9 of 15) described it as stiff, like rubber, small and unevenly warm. On average, participants felt neither stressed (3.26 on a scale of 10) nor anxious (3 on a scale of 10) and were slightly uncertain (4.73 on a scale of 10). The majority of participants successfully identified a hand they were testing as shown above. Since everyone tested each of the two hands twice, the success rate improved linearly for hands that were tested later. By using the cross table method, a weak correlation between participants' feelings in the moment of testing and their success rates can be observed. The majority of participants who recognized two hands wrongly had elevated levels of stress, anxiety and uncertainty at the moment of testing. As a result, the android robot failed the proposed hardware Turing-like test. However, on a positive note, the collected data about the perception of a robot hand by the participants can help improve the human-likeness of robot hands while manufacturing them.
\section{Discussion}

There isn't a known perfect handshake that would satisfy everyone, which makes it difficult to define a ''perfect'' human-like handshake. But a handshake can be modeled to satisfy requirements of a certain social norm. Therefore we propose three ways in which the notion of a Turing test for such hardware interfaces that have complex designs and extreme human-like appearance can be done:
\begin{itemize}
	\item  The first option is to let participants shake robot or human hands and then decide which hand they shook. This option is impossible with the current state of robots because participants would immediately recognize a robot hand. This is evident from the fact that even though we blindfolded participants, they were able to recognize the robot hand just by the sense of touch conveyed by it.
	\item  The second option is to use the robot as an interface for handshaking like a computer is used in the original Turing test. The test would then be whether participants are able to distinguish if the interaction is generated by an algorithm or by a human teleoperating the robot. If they mistake the algorithmic interaction for a teleoperated one then the model can be declared as human-like. This is similar to what is proposed in  \cite{10.1007/978-3-642-14064-8_29},  \cite{article}, \cite{TowardPerceiving2012}
	\item The third option is to compare the trajectories and possibly the gasping force profile generated by different handshake models and of the corresponding human to measure their similarity with human-human interaction trajectories, such as by using/extending the work of \cite{liu2012similarity} or \cite{urain2019generalized}.
	
\end{itemize}


Currently, models for handshaking and other similar interactions have no common criteria to actively judge their acceptance by humans. Therefore, a Turing test for motor intelligence that accurately captures the intricacies of such interactions is required. This way we can compare the quality and results of proposed models in an unbiased manner.


\section{Acknowledgements}
The authors would like to thank the Forum Interdisziplinäre Forschung (Interdisciplinary Research Forum) TU Darmstadt, Dr. Hans Riegel Stiftung, Merck, and the  Förderverein für Marktorientierte Unternehmensführung, Marketing und Personalmanagement e.V. (Association of Supporters of Market-Oriented Management, Marketing, and Human Resource Management) for funding this project.

\bibliographystyle{ACM-Reference-Format}
\bibliography{sample-base}

\end{document}